\documentclass[runningheads]{llncs}
\usepackage{algorithm}
\usepackage{algpseudocode}
\usepackage{array,multirow,adjustbox,graphicx}
\usepackage{subfigure}
\usepackage{amsmath,amssymb} 
\usepackage{color}
\usepackage{tikz}
\usepackage{pgfplots}
\usepackage{hyperref}

\newcommand{\etal}{\emph{et al. }}
\newcommand{\suchas}{\emph{e.g., }}
\newcommand{\thatis}{\emph{i.e., }}

\begin{document}

\title{Water-Filling: An Efficient Algorithm for Digitized Document Shadow Removal\thanks{This research was supported by Hancom Inc.}} 
\titlerunning{Water-Filling} 


\author{Seungjun Jung\inst{1}\orcidID{0000-0001-7150-3746} \and
Muhammad Abul Hasan\index{Hasan, Muhammad}\inst{2}\orcidID{0000-0002-1297-5054} \and
Changick Kim\inst{1}\orcidID{0000-0001-9323-8488}}
%

\authorrunning{Jung et al.} 


\institute{KAIST, Daejeon, Republic of Korea\\
\email{\{seungjun45,changick\}@kaist.ac.kr}\and
University of South Australia, Adelaide, SA, Australia\\
\email{muhammad.hasan@unisa.edu.au}}

\maketitle

\begin{abstract}
In this paper, we propose a novel algorithm to rectify illumination of the digitized documents by eliminating shading artifacts. Firstly, a topographic surface of an input digitized document is created using luminance value of each pixel. Then the shading artifact on the document is estimated by simulating an immersion process. The simulation of the immersion process is modeled using a novel diffusion equation with an iterative update rule. After estimating the shading artifacts, the digitized document is reconstructed using the Lambertian surface model. In order to evaluate the performance of the proposed algorithm, we conduct rigorous experiments on a set of digitized documents which is generated using smartphones under challenging lighting conditions. According to the experimental results, it is found that the proposed method produces promising illumination correction results and outperforms the results of the state-of-the-art methods. \textbf{\textcolor{blue}{[Codes available at :}} \textcolor{green}{\url{https://github.com/seungjun45/Water-Filling}}\textbf{\textcolor{blue}{]}}

\end{abstract}
\section{Introduction}

\begin{figure}[t!]
	\centering
	%
	%
	%
	%
	\includegraphics[width=0.9\linewidth]{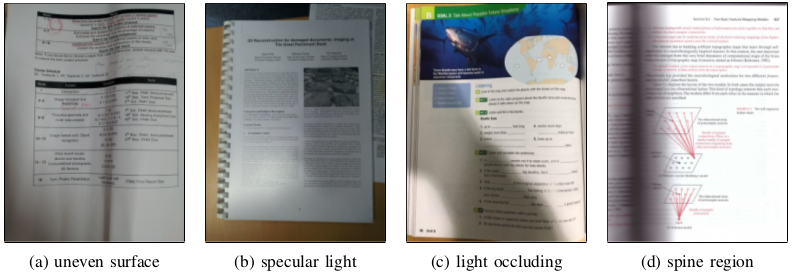}
	
	\caption{Examples of digitized documents having different types of illumination distortion.}
	
	\label{illumination_distortion_examples}
\end{figure}

With the progressive development of built-in cameras in smart handheld devices, people are generating more amount of digital content than ever before. Aside from capturing moments, such powerful cameras are used for capturing digitized copies of printed documents (\suchas important notes, certificates, and visiting cards) for keeping personal records or for sharing the documents with others. This is a popular global trend for its ready availability and relatively easier operational methods than any other digitizing device, \suchas a scanner. Although generating a digitized document using a smart handheld devices is as easy as pointing and shooting the document, the quality of the captured digitized image is often a concern. The paper documents having folds cause to create uneven surfaces and result in illumination distorted documents. Additionally, the specular reflections and hard shadows contribute in rendering poor quality digitized documents. Figures \ref{illumination_distortion_examples}(a)-\ref{illumination_distortion_examples}(c) show three examples of illumination distorted digitized documents captured in different lighting conditions using smartphone's cameras. Even though the scanners are specialized device for producing digitized documents, it also suffer from its own limitations. For example, scanning a page from a thick and bound book suffer from illumination distortions because of the curled area in the spine region \cite{Meng} (see Fig. \ref{illumination_distortion_examples}(d)). Although one can attempt to remove these distortions by pressing the spine of the book harder, such distortions can hardly be eliminated \cite{sfs2,sfs3}.

Correcting illumination distortion of digitized documents is an important task for not only enhancing readability for the readers but also it is crucial for applying optical character recognition (OCR) software on the digitized documents. Various studies in the literature show the importance of rectifying illumination distortion for a significantly better OCR performance \cite{sfs3,OCR1,OCR2,OCR3}. The traditional methods for correcting the illumination distortions of the digitized documents start with estimating the background shades followed by removing the shades using a document surface reconstruction model \cite{Zhang1,Zhang2,Lee,Gato,Oliveira1,Oliveira2,Oliveira3,bdy1,bdy2}. These strategies can be divided into two prominent categories: mask-and-interpolation approaches \cite{Zhang1,Zhang2,Lee,Gato}, and without-mask-interpolation approaches \cite{Oliveira1,Oliveira2,Oliveira3,bdy1,bdy2}.  The former approaches figure out the location of text, and photo regions on the documents using either edge detection or binarization methods, followed by applying a mask that covers the text regions along with its adjacent area on the surface of the documents. Finally, the masked regions are interpolated to correct the illumination of the background of the digitized documents. The latter approaches use color histograms of the document image patches to aggregate the local regions belonging to the background and interpolate the rest of the regions to identify the shaded regions. As removing the shading artifacts using the well-known surface models is a straightforward task, most of the works in the literature concentrated only on extracting the background layers as accurately as possible.

\begin{figure}[t!]
	
	 \centering
	%
	\includegraphics[width=0.9\linewidth]{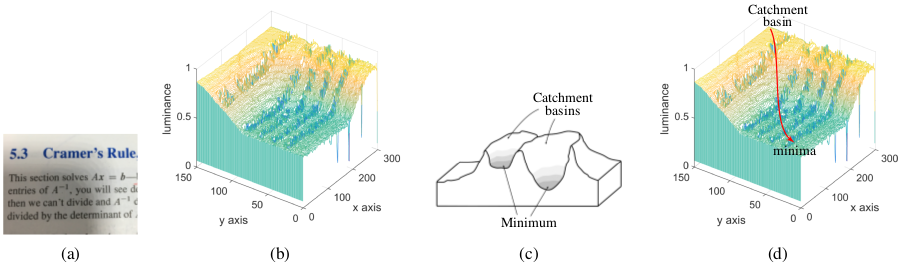}
	
	\caption{A sample digitized image and its topographic surface visualization. (a) Example digitized document, (b) topographic surface representation of (a). (c) An illustration of catchment basins and corresponding minimum. (d) An example of catchment basin and the corresponding minima on a digitized document representing (c).}
	
	\label{simple_simulation}
\end{figure}

In this paper, we propose an efficient illumination correction algorithm for removing shading artifacts from the digitized images. The algorithm starts with constructing a topographic surface using the luminance values of the digitized image pixels irrespective of the contents of the documents (\thatis text and image). Then, a model representing the dynamics of fluids simulated by a diffusion equation is applied on the topographic surface to estimate the shading artifacts on the documents. After estimating the shading artifacts, the digitized documents are reconstructed using the Lambertian surface model to correct the illumination by removing the shades. An example of a topographic surface representation of a digitized image segment is visualized in Fig. \ref{simple_simulation}. In designing the proposed algorithm, we are influenced by the techniques applied in watershed transform \cite{watershed1,watershed2,watershed3}. Accordingly, we named the proposed method called \textit{Water-filling} algorithm for digitized document illumination correction.

The rest of the paper is organized as follows. In Section \ref{sec:RelatedWork}, we present a review of related works in the literature. In Section \ref{sec:proposed}, the proposed Water-filling algorithm for digitized documents illumination correction is described. The experimental results and comparison studies are presented in Section \ref{sec:Experiments}. Finally, the conclusion is presented in Section \ref{sec:Conclusions}.

\section{Related Work}
\label{sec:RelatedWork}
We review the related works in the literature and categorize them into three groups. Each of the groups is discussed in the following subsections.

\subsection{Classical Image Binarization Based Approaches}
One of the traditional approaches for removing shading artifacts from gray-scale digitized documents is binarization technique based methods \cite{Sauvola,bin1,Otsu,Niblack,marginal1,marginal2,bin2}. In \cite{Sauvola,Niblack}, the authors proposed a locally adaptive binarization technique based method which computes a threshold using local information. Bukhari \etal\cite{bin1} improved the method proposed in \cite{Sauvola} by introducing an automatic parameter tuning algorithm using local information. However, it is inevitable that these methods suffer from significant degradation in photographic regions of the digitized documents as such methods convert a gray level pixel to either a black or a white pixel.

\subsection{Parametric Reconstruction Based Approaches}
Illumination distortion on camera generate digitized documents of uneven surface is addressed in the literature using the 3D reconstruction models. Lu \etal \cite{shape2,shape3} proposed a method for removing shading artifacts on digitized documents captured by cameras using the 3D shape reconstruction of the documents. In their method, the 3D shape of the document is reconstructed by fitting the illumination values to a polynomial surface. Meng \etal \cite{meng2012metric} proposed another 3D shape reconstruction based method by isometric mesh construction assuming that the page shape is a general cylindrical surface (GCS). Assuming GCS form for camera captured digitized document images are proven to work well for modeling 3D shape of real documents \cite{kim2015document}. Tian \etal\cite{shape1} proposed another 3D reconstruction based method using text region on the digitized documents. In their method, the perspective distortion of the text region is estimated using text orientation and horizontal text line. Although their proposed methods are effective for correctly identify illumination distortions due to uneven surface of the target document, the method has a limitation in rectifying the shading artifacts on the digitized documents in general (e.g., shadow from light occluding object).

Digitized documents generated by scanners experience relatively less shading artifacts as the scanners use a single light source with an uniform light direction. For such digitized documents, the surface of the document is rendered smooth and constant. Such type of digitized document can be considered to have a parametric surface with various assumptions \cite{Meng,sfs3,scan2}. Such assumptions lead to a straightforward reconstruction of the 3D shapes for digitized documents captured by scanners. The assumption based digitized document illumination correction techniques are not applicable for correcting generalized digitized documents.

\subsection{Background Shading Estimation Based Approaches}

In background shading estimation based methods, a digitized document is assumed to have two separate layers. The background layer--the shading layer which contains illumination distortions, and the foreground layer--the layer which contains the text and images.

\subsubsection{Mask and Interpolation}

In mask and interpolation approach, a mask is created to cover the text region by detecting the text on a digitized document. Zhang \etal \cite{Zhang1,Zhang2} proposed a method to create a mask using the Canny edge detector algorithm followed by using morphological closing operation. This method is effective only for documents which contain the text of a specific size. Documents containing varying font sizes or documents containing large photos result in poor illumination corrected documents. Lee \etal \cite{Lee} proposed another mask and interpolation based method which detects the text and the photo regions with large rectangle masks using the connected edges. This approach is effective for closely captured digitized documents without having any surrounding information. Otherwise, the whole document would be identified as a separate rectangle, hence, that rectangle area would be processed separately. In \cite{Gato}, Gato \etal introduced a method to use the binarized image as a mask. However, the generated masks often fail to cover the photo regions on the digitized documents accurately.  

\subsubsection{Direct Interpolation}

In direct interpolation based approaches, image local regions belonging to the background are detected for illumination correction. Oliveira \etal \cite{Oliveira1,Oliveira2,Oliveira3} assumed that a local region belonging to a digitized document background has a narrow Gaussian shape in its color histogram. With that assumption, initially, the local blocks belonging to the background were identified. Then those blocks were used to estimate the background layer using an interpolation method. Since an image with large size photo has piecewise constant color in the photo region, this method does not work correctly in such cases. In \cite{Bako16}, a shadow map is directly generated using local background and text color estimation. Then the shadow map is applied on the document to rectify the shading artifacts, while document images having irregular shading patterns leads the algorithm to a wrong estimation of the global shadow map. Tsoi \etal \cite{bdy1,bdy2} assumed that boundary region of a digitized document has uniform color. The shading of the interior document was estimated by a linear interpolation of shading using document boundary regions. The method also suffers when shading of the boundary regions has irregular shading patterns.  Fan \etal \cite{fan2007enhancement}  used a watershed-transform segmentation method on the noise-filtered luminance image to segment the background regions. Then, interpolation is applied on the textual regions to estimate illuminant values throughout the image. Since the segmented region with the largest catchment basins is automatically classified as a background region, the performance of the method from this paper is limited when two background regions are separated. 


\section{Water-Filling Algorithm}
\label{sec:proposed}

The proposed digitized document illumination correction algorithm, called Water-Filling algorithm, is described in this section. We start the description with a number of terminologies and relevant observation. The topography of a local area depicts the detail physical construction of land which includes elevation and depression. A \textit{catchment basin} on a topography surface is an extent of a land where the accumulated natural water converges to a meeting point, technically called a \textit{minima}. An illustration of catchment basins and the corresponding minimum is displayed in Fig. \ref{simple_simulation} (c). A digitized document, having shading artifacts, can be represented as a topographic surface, where a depression on the surface is considered as a catchment basin. Figure \ref{simple_simulation} (d) shows an example of a catchment basin appeared on a digitized document due to shading artifact.

\begin{figure}[t]
	\centering
	\includegraphics[width=0.95\linewidth]{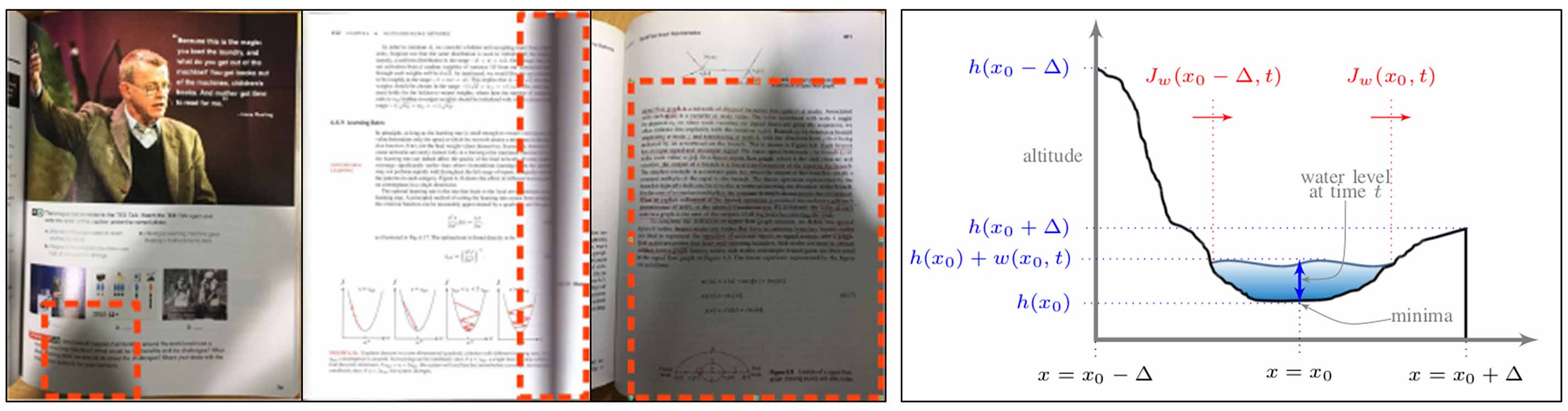}
	
	\caption{(left) Examples of shaded region, marked in orange dashed rectangles, touching the boundary of the images. (right) An one dimensional cartoon topographic model of a catchment basin for illustrating the diffusion equation for document illunination correction.}
	
	\label{figure_shade}
\end{figure}

An illumination distorted region on a digitized document captured by a camera often occurred due to occluding light sources by the camera itself along with the camera holding hand(s) of the user. In case of a scanner, the illumination distorted region often spread from top to bottom or left to right of the document surface. In both device cases, the shaded regions touch the boundary of the digitized documents. Figure \ref{figure_shade} shows a few examples where it can be seen that the shaded region meets the boundary. In the proposed algorithm, we exploit this general observation and summarize that the catchment basins which touch the image boundary correspond to the shading artifacts of the background regions. The other existing catchment basins which do not touch the boundary of the image are in general belonging to the text or photo regions of the documents. We solidify this idea empirically in supplementary material. Based on these observations, let us assume that $D_{I}\subset \mathbb{Z}^{2}$ be a domain of two-dimensional gray-scale image $I$.


In the following sub-sections, the digitized document illumination correction problem is formalized followed by the proposed solution mechanism discussion in detail.

\subsection{Modeling by Diffusion Equation}

In this sub-section, we model a filling mechanism of catchment basins using water. We use two different methods to simulate this water-filling task: \textit{incrementally filling of catchment basins} method, and \textit{flood-and-effuse} method. 
Ultimately, we link these two methods to provide the final version of the water-filling method.

Before modeling each method using diffusion equation, we need to set a few constraints on a few variables. Let $h$ be the altitude of the topographic surface, $w (x, t)$ be the water level on the topographic surface $x$ at a point of time $t$. Figure \ref{figure_shade} illustrates an one-dimensional cartoon model of a catchment basin. Our objective is to restore the background of the digitized document affecting the structure of the document as low as possible. In order to meet  the objective, we set two constraints  on $w()$ given as follows.

\begin{align}
\label{constraint:level}
w (x,t) \geq 0, & \quad \forall t,\ \forall x \in D_{I}, \\
\label{constraint:boundary}
w(x,t) = 0, & \quad \forall t,\ \forall x \in \partial D_{I}.
\end{align}
Equation (\ref{constraint:level}) limits water level to non-negative value since water either be stored or flow out. Equation (\ref{constraint:boundary}) describes the drop of water at the image boundary so that only the catchment basins at the interior region of the image can be filled. Under these constraints, the dynamics of water can be designed with the following diffusion equation.

\begin{equation}{\label{PDE:1}}
\begin{split}
\frac{\partial w(x,t)}{\partial t}\bigg |_{x=x_{0}} &= -\nabla \cdot \mathbf{J_{w}} (x,t)\big |_{x=x_{0}} =-\frac{\partial J_{w} (x,t)}{\partial x}\bigg |_{x=x_{0}},
\end{split}
\end{equation}
where $\mathbf{J_{w}}$ is the flux of the diffusing water. Equation (\ref{PDE:1}) states that a change in water level in any part of the structure is due to inflow and outflow of water into and out of that part of the structure.

\subsubsection{Incremental Filling of Catchment Basins}

Let $G( x_{0},t)$ be the overall altitude of water at a location $x_0$ at time $t$, which can be written as $G( x_{0},t) = h( x_{0}) + w(x_{0},t)$. Assuming that there is a continuous water supply source from the peak $\hat{h}$, the total flow of the water is directly proportional to its relative height at $x$. Formally, we write it as follows.

\begin{equation}
J_{w}(x_{0},t) \propto G( x_{0},t)-G(x_{0}+\Delta,t).
\label{PDE:2}
\end{equation}
Similarly, for the 2D case of the flow dynamics, the partial derivative of $w()$ with respect to time $t$ at $(x_{0},y_{0})$ is proportional to the relative difference of $G()$ at $(x_{0},y_{0})$ and its neighborhood, where $(x, y)$ represents a point on the 2D topographic surface. Following Equation (\ref{PDE:2}), the 2D iterative update formula for $w()$ is given in Equation (\ref{PDE:3}).


\begin{equation}{\label{PDE:3}}
\begin{split}
w(x,y,t+\Delta) = & \eta \cdot \big\{G( x+\Delta,y,t) +G( x-\Delta,y,t) +G(x,y+\Delta,t) \\ 
& + G( x,y-\Delta,t)-4\cdot G( x,y,t)  \big\} +w( x,y,t),
\end{split}
\end{equation}
where $\eta$ is a hyperparameter which decides the speed of the process. The $\eta$ has to be chosen carefully as an inappropriate value of $\eta$ might lead to slow convergence or divergence of the water level. Since we use $G()$ values from four-neighborhood at point $(x,y)$, the proper $\eta$ value is less than or equal to $0.25$. The value of $\eta$ greater than $0.25$ cause flooding in the whole structure. The value of $G()$ after convergence is used to estimate the background layer of an image. The $G()$ is converged at time $t$, if $G(x,y,t) - G(x,y,t-\Delta) < \delta$, $\forall x, \forall y$, where $\delta$ should be a small value. In this paper, we have set $\delta = 0.01$. 

\begin{figure}[t!]
    \centering

    \includegraphics[width=1\linewidth]{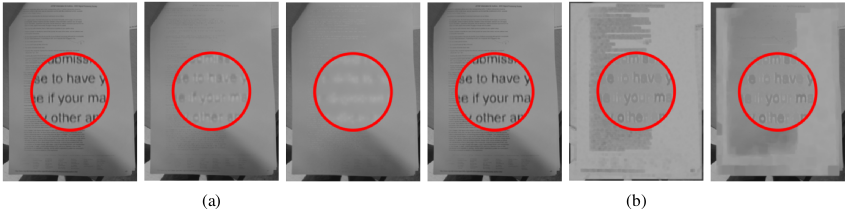}
    
    \caption{Estimation of background layer $G(x,y,t)$ using our iterative water-filling approach (left to right: $t=0, 50, 200$): (a) incremental filling of catchment basins and (b) flood-and-effuse method. The red color circles show 10 times magnified content at the center of the documents for better visual understanding.}
    
\label{simulation}
\end{figure}

\subsubsection{Flood and Effuse}

In this method, the diffusion equation can be decomposed into two independent processes: water effusion, and flood. For the effusion process, we consider the dynamics of water flow on the topographic surface without any external water supply. We only consider the effusion of water through the image border. Also, the amount of water at each location never be increased in order to further boost the speed of effusion process. The formulation of this idea for the 1D case is straightforward from Equation (\ref{PDE:1}) and (\ref{PDE:2}) is given as follows. 


\begin{equation}{\label{effuse}}
\begin{split}
w_{\phi}(x_{0},t) & \propto \min \{-G(x_{0},t) + G(x_{0}+\Delta,t),0\}  \\
& + \min\{-G(x_{0},t)+G( x_{0}-\Delta,t),0 \}. 
\end{split}
\end{equation}
The effusive term $w_{\phi}(x_{0},t)$ has non-positive value and represents the amount of water to be effused at each location. Conversely, in the flooding process $w_{\psi}( x_{0},t)$, the model be immersed to the same altitude for few initial moments. 

\begin{equation}
w_{\psi}( x_{0},t) = ( \hat{h}-G( x_{0},t))\cdot e^{-t},
\label{flood}
\end{equation}
where $\hat{h}={\max} \ h(x), \forall x$. The partial derivative of the overall water level $w()$  with respect to time $t$ is the sum of $w_{\phi}$ and $w_{\psi}$. 
The iterative update formula for $w()$ in the 2D case is as follows.

\begin{equation}{\label{PDE_4}}
\begin{split}
w(x,y,t&+\Delta) = ( \hat{h} -G(x,y,t)) \cdot e^{-t} + \eta \cdot \{ \min \{G(x + \Delta, y, t) -G(x,y,t), 0 \} \\
& + \min \{G(x-\Delta,y,t)-G( x,y,t),0\} + \min \{G(x,y+\Delta,t)-G(x,y,t),0\}  \\
& + \min \{G(x,y-\Delta,t)-G(x,y,t),0  \} \} +w(x,y,t). 
\end{split}
\end{equation}

Once $G()$ converges, it can be used to reconstruct photometrically correct version of the digitized documnet. In Lambertian surface model, the illumination value of an image is equal to the product of foreground text layer and background shading layer \cite{Zhang1}. Given the background layer $I_{b}$ and foreground layer $I_{f}$, the final photometrically correct image, $I_{r}$, can be computed as follows.

\begin{equation}
I_{r}\left(x,y \right )=\ell \cdot I_{f}=\ell \cdot \frac{I\left(x,y \right )}{I_{b}\left(x,y \right )} \approx \underset{t \rightarrow \infty }{\lim} \frac{I\left(x,y \right )}{G\left(x,y,t \right )} \cdot \ell,
\label{Lambertian}
\end{equation}
where $\ell \in [ 0,1]$ for tuning the brightness of the output image. A few example of simulation results of the above two methods are shown in Fig. \ref{simulation}. It is notable that the results from the flood method show the desired dynamics precisely -- once the overall image is filled with water, it gradually effuses water at the image boundary.


Although the incremental filling method removes shading artifacts quite well, the photo regions are not clearly restored because the filling process at the large catchment basins are too slow. Meanwhile, it is observed that the overall image is reconstructed correctly with the flood-and-effuse method.

%

\subsubsection{Boosting with Downsampling}
The time complexity of the proposed algorithm is $\mathcal{O}(I_{width} \times I_{height} \times t)$. Since both of the Equations (\ref{PDE:3}) and (\ref{PDE_4}) are pixel-wise operations, the processing time increases as the image size gets larger, which hinders these methods from being used in real-time applications. Although applying either equation after downsampling the input document size at a rate of $k_s$ can boost the processing time significantly, we observe that background shadings on small catchment basins may not be estimated correctly due to aliasing effect. However, shading artifacts in a small region do not have a drastic effect on the overall digitized document illumination correction. Thus, we apply the flood-and-effuse method on the downsampled input document only to attain the rough approximation quickly. After estimating the background artifacts, using bicubic interpolation the background layer is approximated to the original size. Then, the incremental filling method is used to reconstruct the details of the rectified digitized document. In the following section, we show that this combined scheme produces reliable results in a shorter time with acceptable peak signal-to-noise ratio (PSNR). The pseudocode of the proposed algorithm is given in Algorithm \ref{Water_filling}.

\begin{algorithm}[h!]
\footnotesize
\caption{Water-Filling Algorithm}\label{Water_filling}

\begin{algorithmic}[1]

\Procedure{Water-filling}{$Img$, $k_s$}
   \State $[Y,Cb,Cr] \gets \text{RGB\_to\_YCbCr} ~(Img)$
   
   \State $\overline{Y} \gets \text{Downsample}(Y, k_{s})$  \Comment{Sampling rate of $k_{s}$}
   
   \State $\overline{w} \gets \text{Flood}(\overline{Y})$ \Comment{Rough sketching} 
   \State $\overline{G} \gets \overline{w} + \overline{Y}$
   \State $G \gets \text{Upscale}(\overline{G},k_{s})$ \Comment{Bicubic interpolation}
   \State $w \gets \text{Incremental}(G)$ \Comment{Detail reconstruction}
   \State $G \gets w+G$ \Comment{Final background}
   \State $Y \gets \ell \times Y/G$ \Comment{$\ell$ is a brightness factor}
   \State $Img_{c} \gets \text{YCbCr\_to\_RGB}([Y,Cb,Cr])$
   \State \textbf{return} $Img_{c}$
\EndProcedure
\end{algorithmic}
\end{algorithm}


\section{Experiments and Performance Evaluation}
\label{sec:Experiments}

\subsection{Datasets}
Since there is no publicly available digitized document illumination distortion dataset, we have created our own to compare the performance of our proposed method with the state-of-the-art methods. Primarily our dataset consists of $159$ illumination distorted digitized documents, among them $109$ images are captured using two smartphone's cameras and the rest $50$ images are captured using a scanner. The size of camera captured images are $3264 \times 2448$, and the scanned images are $3455 \times 2464$ ($72$ dpi). The digitized document captured by cameras are generated under different lighting conditions. For PSNR comparison, we captured $87$ ground-truth images along with its illumination distorted digitized images. In order to capture ground-truth images, a camera stand is used in a well-lit room to capture a ground-truth photo followed by capturing another image by adding shades on the document surface by occluding light sources intentionally. We present the benchmark dataset in our supplementary material. We further collected  $22$ business card images from a publicly available dataset \cite{namecard} for OCR edit-distance comparison. The proposed algorithm is implemented using Microsoft C++ development environment. For experiments, we use a workstation having an Intel Corei7-4790k, $64$ bits CPU with $16$ GB memory and $128$ SSD storage device.

\subsection{Methods for Comparison}
The performances of the proposed algorithm have been compared with several state-of-the-art methods in the literature. For comparing the performance on the digitized documents captured by smartphones, we select a binarization method (Sauvola \etal \cite{Sauvola}), a mask and interpolation method (Zhang \etal \cite{Zhang1}), two direct interpolation methods (Oliveira \etal \cite{Oliveira3}, Bako \etal \cite{Bako16} and Kligler \etal \cite{kligler2018document}), and a rolling-ball based method (Sternberg \etal \cite{Sternberg}). The rolling ball based method estimates a local background value for every pixel by taking an average over a large ball around the pixel. For comparing the performance on digitized documents captured by scanner devices, we select the methods proposed in \cite{Sauvola}, \cite{Sternberg}, \cite{Oliveira3}, \cite{Meng} and \cite{Zhang1}. 


\subsection{Parameters Tuning}
\begin{figure}[t!]
    \centering

    \includegraphics[width=0.95\linewidth]{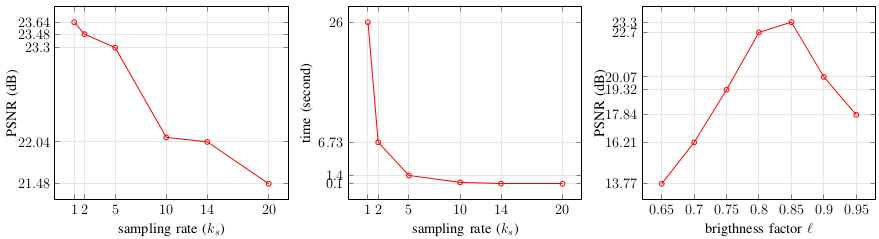}
    
    \caption{(a) The trade-off between the sampling rate and PSNR, and (b) the trade-off between the sampling rate and average time taken to correct illumination distortion of a digitized document. In this experiment, we set brightness factor $\ell = 0.85$. (c) The trade-off between the brightness $\ell$ and PSNR. In this experiment we set the sampling rate $k_s = 5$.}
    
\label{fig_subsampling_rate}
\end{figure}

There are two hyperparameters to be set in our algorithm for generating optimum results, namely sampling rate $k_s$ and the brightness factor $\ell$. In order to do that, first of all, we conduct an experiment using the first 20 distorted images along with their ground-truth images. We need to set an acceptable $k_s$ to correct shading artifacts fast by meeting our objective, that is we want estimate shadow as clearly as possible. Too large $k_s$ makes the algorithm preform faster at a cost of poor quality output. In order to produce acceptable results using the proposed algorithm, we need to look after two crucial components, the PSNR in the corrected images in comparison to the ground-truths, and the average elapsed time to accomplish illumination rectification. Accordingly, we conduct two experiments for selecting an acceptable $k_s$ which are reported in Fig. \ref{fig_subsampling_rate}. It is obvious that, by selecting a larger $k_s$ makes the algorithm perform faster by a small margin. However, that would cost the algorithm performing poorly by decreasing significantly large amount of PSNR. Accordingly, we select $k_s = 5$ for the rest of the experiments as the default sampling rate.

The brightness of the output image is also an important factor. The output document should be comfortable to the human eyes--too bright and too dark both are undesirable. To determine the correct brightness factor, we conduct another experiment to measure PSNR for changing brightness factor $\ell$. Figure \ref{fig_subsampling_rate} (c) shows the experiment results. As it can be seen, for brightness factor $\ell = 0.85$, we achieve optimum results. Based on this experiment, we set $\ell = 0.85$ for the rest of the experiments to be done.

\begin{figure*}[t!]
    \centering
	\subfigure[input]{\includegraphics[width=0.17\linewidth]{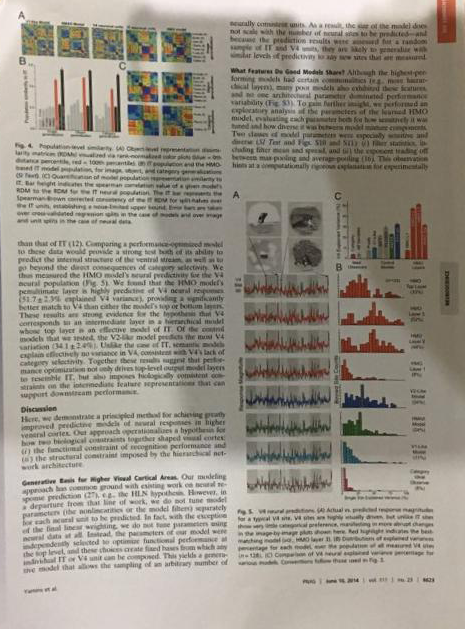}}
	\subfigure[using \cite{Sauvola}]{\includegraphics[width=0.17\linewidth]{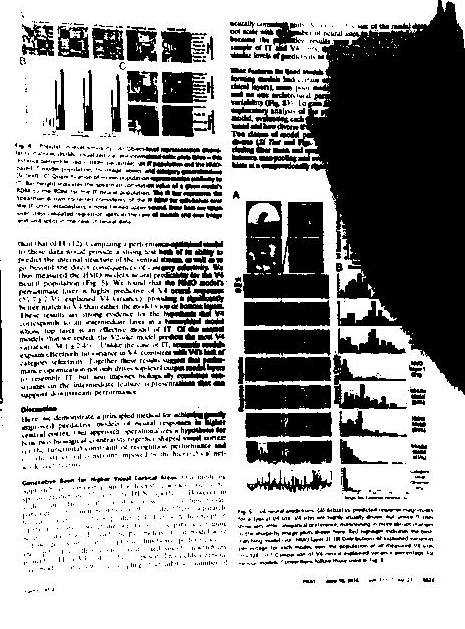}}
	\subfigure[using \cite{Zhang1}]{\includegraphics[width=0.17\linewidth]{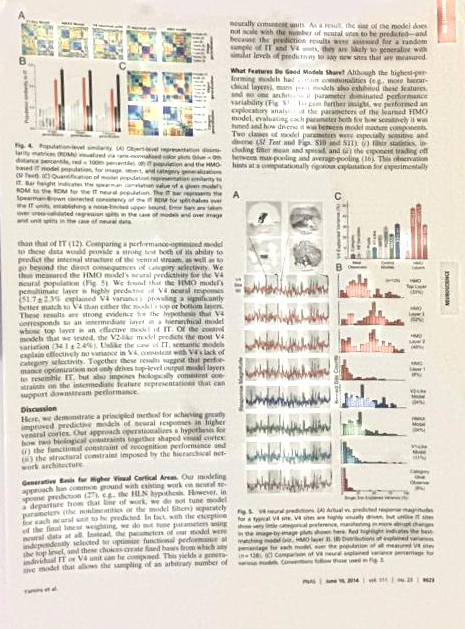}}
	\subfigure[using \cite{Sternberg}]{\includegraphics[width=0.17\linewidth]{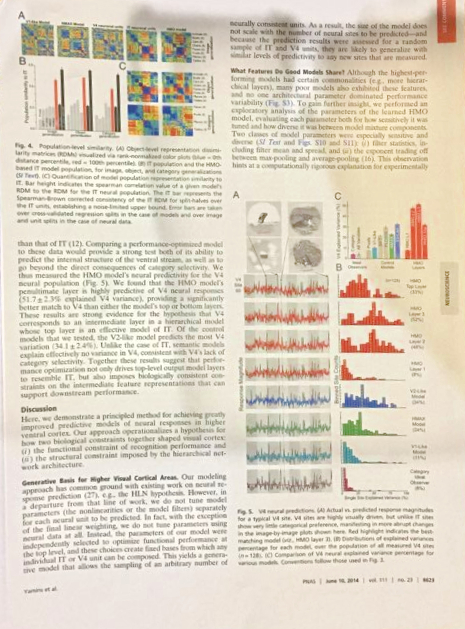}}
	\subfigure[using \cite{Oliveira3}]{\includegraphics[width=0.17\linewidth]{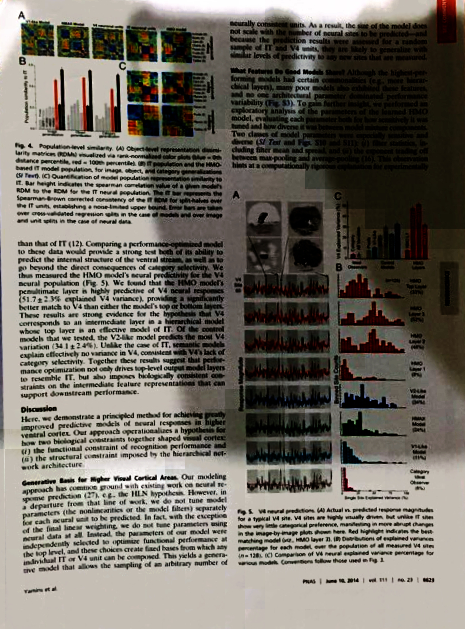}}
	\subfigure[using \cite{Bako16}]{\includegraphics[width=0.17\linewidth]{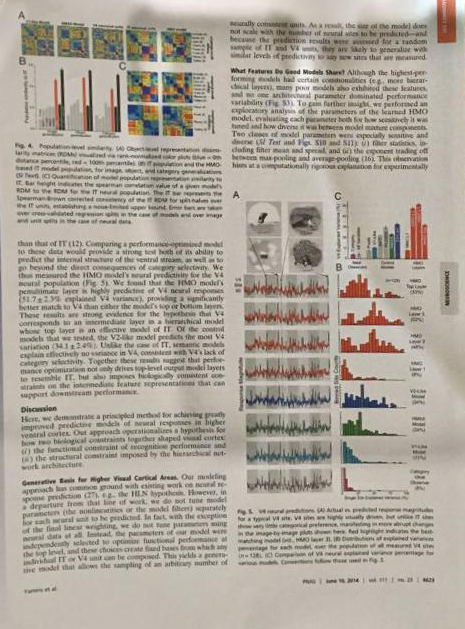}}
	\subfigure[using \cite{kligler2018document}]{\includegraphics[width=0.17\linewidth]{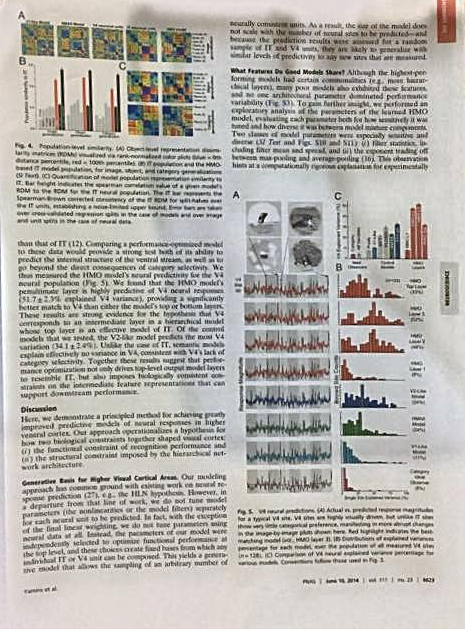}}
	\subfigure[ours]{\includegraphics[width=0.17\linewidth]{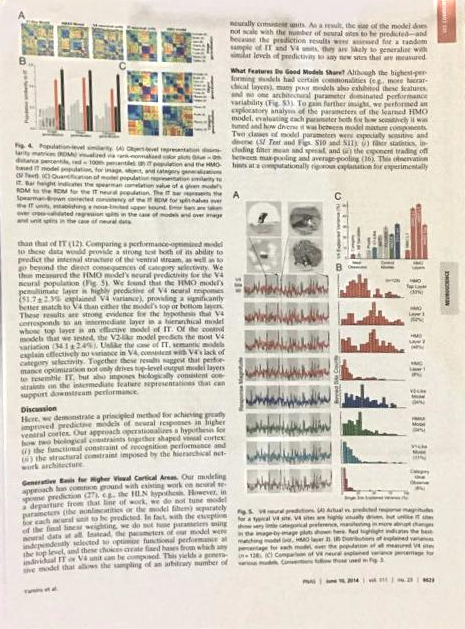}}
	\subfigure[gt]{\includegraphics[width=0.17\linewidth]{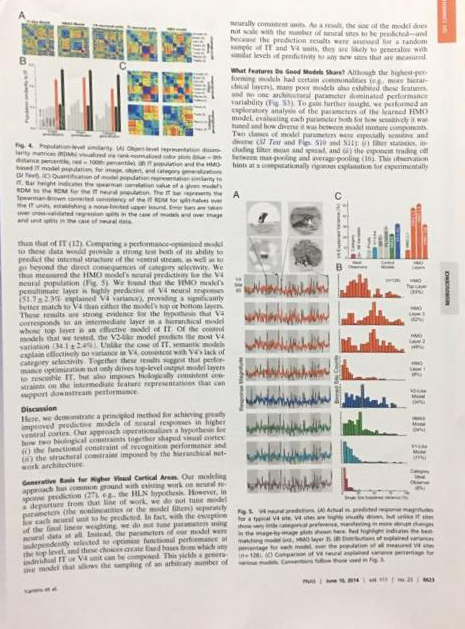}}

    \caption{An example of illumination distorted digitized documents (generated using a smartphone camera mounted on a camera stand) and its illumination corrected results using several methods.}
    
\label{overall1}
\end{figure*}

%

\subsection{Results Analysis}

After tuning the hyperparameters of the proposed method, we conduct a visual comparison of the illumination corrected results of ours method and that of a number of existing methods. Figure \ref{overall1} shows an example digitized documents along with the ground-truth and the corresponding illumination corrected results performed by several methods. 
For \cite{Oliveira3,Bako16,kligler2018document}, false detection of background region results in unclear rectification of illumination distortions along the shadow borders. Moreover, their method often adds additional noises in the text and image regions which cause in destroying the structures of the documents. Another important concern of method \cite{Oliveira3} is that the photo regions on the documents render into dark patches. For the image region on the document, the method \cite{Zhang1} fails to preserve the structure and color, in fact, it ruins the visual quality in those regions.

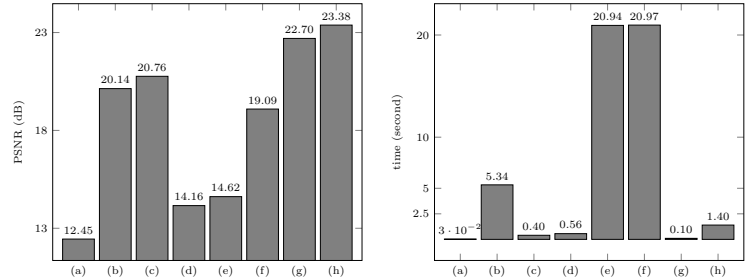
\begin{figure}[!htbp]

\centering
\subfigure{
\begin{tikzpicture}[scale=0.6]
\begin{axis}[
ylabel=PSNR (dB),
ylabel near ticks,
symbolic x coords={(a), (b), (c), (d), (e), (f), (g), (h)},
font=\scriptsize,
nodes near coords,
nodes near coords align={vertical},
every node near coord/.append style={
/pgf/number format/fixed zerofill,
/pgf/number format/precision=2
},
bar width=20pt,
xtick=data, ytick = {23.00, 18.00, 13.00, 8.00}
]
\addplot[ybar,fill=gray] coordinates {

((a), 12.4481)
((b), 20.1352)
((c), 20.7643)
((d), 14.1576)
((e), 14.6161)
((f), 19.091)
((g), 22.70)
((h), 23.3807)
};

\end{axis}
\end{tikzpicture}
}
\subfigure{
\begin{tikzpicture}[scale=0.6]
        \begin{axis}[
        ylabel=time (second),
        ylabel near ticks,
            symbolic x coords={(a), (b), (c), (d), (e), (f), (g), (h)},
            font=\scriptsize,
            nodes near coords,
     nodes near coords align={vertical},
     every node near coord/.append style={
    /pgf/number format/fixed zerofill,
    /pgf/number format/precision=2, 
},
            bar width=20pt,
            xtick=data, ytick = {20.0, 10.0, 5.0, 2.5}
          ]
            \addplot[ybar,fill=gray] coordinates {
((a), 0.03)
((b), 5.335)
((c), 0.4)                
                ((d), 0.56)
                ((e), 20.94)
                ((f), 20.97)
                ((g), 0.10)
                ((h), 1.40)
            };
            
        \end{axis}
    \end{tikzpicture}
}
\caption{(left) Ground-truth comparison in terms of average PSNR and (right) average processing time comparison (a) using \cite{Sauvola}, (b) using \cite{Zhang1}, (c) using \cite{Sternberg}, (d) using \cite{Oliveira3}, (e) using  \cite{Bako16}, (f) using \cite{kligler2018document}, (g) ours (using $k_s = 14$) and (h) ours (using $k_s = 5$).}
\label{fig_psnr_comparison}
\end{figure}

%
%

In order to have a fairer comparison, the methods are applied to the ground-truth data to compare the PSNR. Figure \ref{fig_psnr_comparison} summarizes the performances of the methods on ground-truth data where it can be seen that the proposed method outperforms the other compared methods in terms of PSNR. The closest PSNR to our method is achieved by method \cite{Sternberg} with the parameter ball size $=50$. However, the PSNR difference between these two method is quite significant and our method is proved to be superior in producing better illumination corrected digitized documents.

The average time requires to produce a rectified document is another important criteria to be measured. From the experimental results shown in Fig. \ref{fig_psnr_comparison}, the method \cite{Sauvola} take least amount of average time ($3 \times 10^{-2}$ sec.) to accomplish the task, while the method \cite{Bako16} takes the highest amount of average time to correct illumination of a document. Among the rest of the reported methods, as outputs produced by method \cite{Sauvola} are severely distorted, we can disregard their performance. Among the rest, our proposed method, using sampling rate $k_s = 5$, takes $1.4$ seconds to accomplish the task. Although the elapsed time ($1.4$ second) for this setting is higher than the elapsed time of the other remaining methods, by setting the sampling rate $k_s = 14$, our method can accomplish the task in $0.1$ second. In that case, the documents have to suffer from more noise (PSNR 22.04, see Fig. \ref{fig_subsampling_rate}) than that of sampling rate $k_s = 5$. However, the PSNR level is still better than that of methods \cite{Oliveira3}, \cite{Zhang1}, and \cite{Sternberg} (see Fig. \ref{fig_psnr_comparison}). That means, even if a coarse sampling rate is chosen for the proposed algorithm, the output quality of the documents is still better than the other compared methods.

\begin{figure}[t!]
    \centering
    \includegraphics[width=0.9\linewidth]{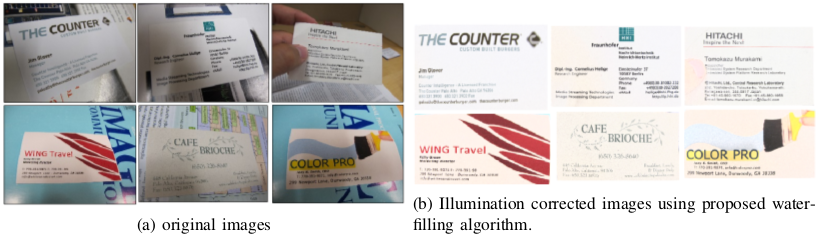}
    
    \caption{Examples of business cards chosen from \cite{namecard} and the corresponding illumination corrected, and geometrically de-warped results after applying the proposed algorithm.}
    
\label{namecard1}
\end{figure}

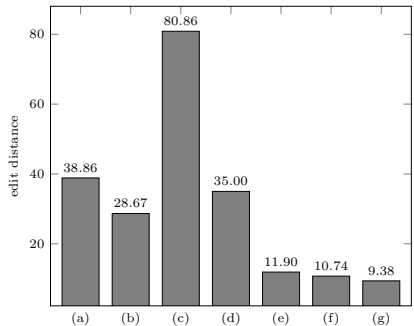
\begin{figure}[!htbp]
\centering
\begin{tikzpicture}[scale=0.7]
        \begin{axis}[
        	ylabel=edit distance,
        	ylabel near ticks,
            symbolic x coords={(a), (b), (c), (d), (e), (f), (g)},
            font=\scriptsize,
            nodes near coords,
    		 nodes near coords align={vertical},
    		 every node near coord/.append style={
    		/pgf/number format/fixed zerofill,
    		/pgf/number format/precision=2
			},
            bar width=20pt,
            xtick=data, ytick = {80.0, 60.0, 40.0, 20.0}
          ]
            \addplot[ybar,fill=gray] coordinates {
				((a), 38.86)
				((b), 28.67)
				((c), 80.86)                
                ((d), 35.00)
                ((e), 11.90)
                ((f), 10.74)
                ((g), 9.38)
            };
            
        \end{axis}
    \end{tikzpicture}

\caption{Average edit-distance measure (after applying OCR) comparison (a) using original images, (b) using \cite{Sternberg}, (c) using \cite{Zhang1}, (d) using \cite{Oliveira3}, (e) using  \cite{Bako16}, (g) using \cite{kligler2018document}, and (f) ours.}

\label{fig_edit_dist}
\end{figure}

While our method shows superior performance over other compared methods presented in the previous sub-section, it is also important to examine whether the structure of characters is well preserved after illumination correction. To this end, we measure the OCR performance for each illumination corrected output, as an indication of readability of the methods. 
Before applying OCR, the corrected digitized documents are geometrically de-warped using inverse perspective projective mapping by manually selecting four corner points of each document. 
Then, we apply an online OCR implementation of \cite{smith2009hybrid} to recognize texts on the documents. In order to compare the performance of our algorithm with the state-of-the-art methods, we measure the average edit-distance between the OCR results and the original text written on the cards. The edit-distance measure is reported in Fig. \ref{fig_edit_dist}. As it can be seen, the average edit-distance measure of the text on the illumination corrected digitized business cards by our algorithm is $9.38$ while the nearest edit-distance measure is performed by method \cite{kligler2018document} having the performance score $10.74$. Besides, the proposed method outperforms the other reported method by a significantly large margin. Based on the experiments using OCR results, we strongly claim that the proposed method is not only capable of correcting illumination of the digitized documents but also keep the structure of the documents well preserved, which helps the user for experiencing better readability using the digitized document corrected by our proposed algorithm. We provide further experimental results (\suchas{scanner}) in supplementary material.

\section{Conclusions}
\label{sec:Conclusions}

In this paper, we have proposed a novel algorithm, called water-filling algorithm, for illumination correction of shaded digitized documents. The proposed water-filling algorithm is based on simulating the immersion process of a topographic surface using water. We have implemented this idea using a diffusion equation. To the best of our knowledge, no previous method used diffusion equation based approach for digitized document illumination correction. Based on the experimental results, it is found that the proposed method can outperform the performance of the previous methods under various challenging lighting conditions. The limitation of the proposed algorithm is that any text or photo connected to image border is regarded as a shadow, and eliminated after the processing. However, we have observed that this is not a critical problem since the salient text or photo is usually not connected to image borders in a document image. We also need to mention that our method often produces unsatisfactory results for eliminating specular lights in an image, because once the original luminance value of a point in the foreground layer is significantly damaged due to overexposure, it is hard to reconstruct the point using the Lambertian surface model even if we precisely estimated the background layer.  These limitations will be addressed in our future works.

\bibliographystyle{splncs04}
\bibliography{egbib_QC}

\end{document}